\pdfoutput=1

\documentclass[11pt,table]{article}
\usepackage{tikz}
\usepackage{authblk}
\usepackage{hyperref}
\def\checkmark{\tikz\fill[scale=0.4](0,.35) -- (.25,0) -- (1,.7) -- (.25,.15) -- cycle;}
\usepackage[]{acl}
\usepackage{multirow}
\usepackage{graphicx}
\usepackage{amssymb}
\usepackage{times}
\usepackage{url}
\usepackage{latexsym}
\usepackage{booktabs}
\usepackage[T1]{fontenc}

\usepackage[utf8]{inputenc}

\usepackage{microtype}
\usepackage{xcolor}
\usepackage{todonotes}
\usepackage{multirow}

\newcommand{\injy}[1]{{\color{black}#1}}

\title{UU-Tax at SemEval-2022 Task 3: Improving the generalizability of language models for taxonomy classification through data augmentation}

\author[1,2]{Injy Sarhan}
\author[1]{Pablo Mosteiro}
\author[1,3,4]{Marco Spruit}
\affil[1]{Department of Information and Computing Sciences, Utrecht University, The Netherlands. } 
\affil[2]{Arab Academy for Science, Technology, and Maritime Transport, Egypt. } 

\affil[3]{Department of Public Health and Primary Care, Leiden University Medical Center, \protect\\ The Netherlands.}
\affil[4]{Leiden Institute of Advanced Computer Science, Leiden University, The Netherlands.\protect\\  \texttt{i.a.a.sarhan@uu.nl} ,  \texttt{p.mosteiro@uu.nl}\texttt{ m.r.spruit@lumc.nl} \protect\\ }

\begin{document}

\maketitle
\begin{abstract}
This paper presents our strategy to address the SemEval-2022 Task 3 PreTENS: Presupposed Taxonomies Evaluating Neural Network Semantics. The goal of the task is to identify if a sentence is deemed acceptable or not, depending on the taxonomic relationship that holds between a noun pair contained in the sentence. For sub-task 1---binary classification---we propose an effective way to enhance the robustness and the generalizability of language models for better classification on this downstream task. We design a two-stage fine-tuning procedure on the ELECTRA language model using data augmentation techniques. Rigorous experiments are carried out using multi-task learning and data-enriched fine-tuning. Experimental results demonstrate that our proposed model, UU-Tax, is indeed able to generalize well for our downstream task.  For sub-task 2---regression---we propose a simple classifier that trains on features obtained from Universal Sentence Encoder (USE). In addition to describing the submitted systems, we discuss other experiments that employ pre-trained language models and data augmentation techniques. For both sub-tasks, we perform error analysis to further understand the behaviour of the proposed models.  We achieved a global F1\(_{\rm Binary}\) score of 91.25\% in sub-task 1 and a rho score of 0.221 in sub-task 2.\footnote{Our implementation of UU-Tax is publicly available at \url{https://github.com/IS5882/UU-TAX}.}
\end{abstract}


\section{Introduction}
\label{sec:intro} 

Predicting the semantic relationship between words in a sentence is essential for Natural Language Processing (NLP) tasks.
Deep neural language models accomplish outstanding results in multiple tasks involving semantics evaluation. The question posed by the shared task Presupposed Taxonomies: Evaluating Neural Network Semantics (PreTENS) is whether neural models can detect the taxonomic relationship between nouns, especially in scenarios where the pattern and/or the set of nouns in the sentence is previously unseen~\cite{zamparelli-etal-PreTENS22}. Sub-task~1 is a simpler classification task, while sub-task 2 is a more complex regression task. Both sub-tasks involve datasets in English, French and Italian. For each sub-task, teams are permitted three submissions. For each submission, the score is averaged over the three languages. The highest score from the three submissions is reported.

We propose a series of models based on pre-trained language models. We enhance the provided datasets using state-of-the-art data augmentation tools, and further increase the dataset size by employing translations. The aim of both steps is to create slightly modified versions of the sentences, such that the model can learn alternative forms of nouns and patterns.

For the classification task (sub-task 1), we obtained the 3$^{\rm rd}$ place, with an F1\(_{\rm Binary}\) score of 91.25\% averaged over the three languages. For the regression task (sub-task 2), we obtained the 5$^{\rm th}$ place, with a Spearman's correlation coefficient $\rho$ of 0.221 averaged over the three languages. Sub-task 2 is markedly more difficult than sub-task 1 due to sentences that can be ambiguous, such as \emph{I like dogs, but not chihuahuas}; some humans will judge this sentence as acceptable, while some will not. 
We attempt to solve both tasks by employing data augmentation techniques in order to help the models understand variations in text. Our main contributions are: \textit{(i)} we devise a special development-validation split to emulate the real situation in which the model must face new words and patterns, and \textit{(ii)} we combine various data augmentation tools to allow the models to learn from various versions of the training dataset.

In Section~\ref{sec:related_work} we present the task details and some of the related work that was done previously. In Section~\ref{sec:System} we motivate our choice of models. The experiments we performed are in Section~\ref{sec:experimental_setup}. Results and conclusions are presented in Sections~\ref{sec:results} and~\ref{sec:conclusion}.


\section{Background}
\label{sec:related_work}
For the present task, we are provided with a list of sentences following a set of \emph{patterns}, all of which have two slots for noun phrases. One such sentence might be:
\(\emph{I don't like beer, a special kind of drink}\).
The pattern corresponding to this sentence would be:
\emph{I don't like {\rm [blank]}, a special kind of {\rm [blank]}}.
Sentences are labeled according to whether the taxonomic relation between the two nouns makes sense. In sub-task 1, labels are binary; a sentence such as that shown above has a label of 1, while this sentence would have a label of 0:
\(\emph{I like huskies, and dogs too}\).
In sub-task 2, labels are continuous, ranging from 1 to 7; these scores are based on a seven-point Likert scale, judged by humans via crowdsourcing. The same dataset is presented in English, Italian and French. For sub-task 1, the training and test sets consist of 5\,838 and 14\,556 sentences, respectively; for sub-task 2, the training and test sets consist of 524 and 1\,009 sentences, respectively. 

There are two challenges to this dataset: \emph{(i)} The test dataset is much bigger than the training dataset, and \emph{(ii)} There are unseen patterns and noun pairs in the test set.
The combination of these hampers the ability of machine learning (ML) models trained on the training set to generalize well to the test set. Indeed, that is the aim of this task: to evaluate the ability of language models to generalize to new data when it comes to inferring taxonomies.

One way to conceptualize the PreTENS task is to reformulate it as a taxonomy extraction task with pattern classification and distributed word representations. For a given sentence, extract the noun pair and the pattern from the sentence, and then determine if the taxonomic relation between the nouns matches the relations allowed by the pattern.  
This formulation is motivated by previous work in taxonomy construction that relied on various approaches ranging from pattern-based methods and syntactic features to word embeddings \cite{chineseunsupervised,luu2016learning,hearstrevisted}. 
As promising as this approach sounds for PreTENS, it involves manual labeling of the noun-pair taxonomic relations in the training set, as we are not allowed to use resources such as WordNet \cite{wordNet} or BabelNet \cite{babelnet}.

A different approach is to tackle PreTENS as a cross-over task between extraction of lexico-semantic relations and commonsense validation. There have been SemEval tasks to extract and identify taxonomic relationships between given terms (SemEval-2016 task 13) \cite{bordea2016semeval}, and to validate sentences for commonsense (SemEval-2020 task 4,  sub-task A) \cite{wang2020semeval}. The aim of the common-sense validation task is to identify which of two natural language statements with similar wordings makes sense.

In the SemEval-2016 task 13, approaches related to extracting hypernym-hyponym relations to construct a taxonomy involved both pattern-based methods and distributional methods. TAXI relied on extracting Hearst-style lexico-syntactic patterns by first crawling domain-specific corpora based on the terminology of the target domain and later using substring matching to extract candidate hypernym-hyponym relations \cite{panchenko2016taxi}. Another team designed a semi-supervised model based on the hypothesis that hypernyms may be induced by adding a vector offset to the corresponding hyponym word embedding \cite{pocostales2016nuig}.

Participants in the SemEval 2020 commonsense validation task had an advantage over PreTENS participants: they were allowed to integrate taxonomic information from external resources such as ConceptNet \cite{wang2020semeval}, which eased the process of fine-tuning the language models on the down-stream task. As an example, the CN-HIT-IT.NLP team \cite{zhang2020cn} and ECNU-SenseMaker \cite{zhao2020ecnu} both used a variant of K-BERT \cite{liu2020k} with additional data; the former injects relevant triples from ConceptNet to the language model, while the later also uses ConceptNet’s unstructured text to pre-train the language model. Other systems relied on ensemble models consisting of different language models such as RoBERTa and XLNet \cite{liu2020qiaoning,altiti2020just}.

In Section~\ref{sec:System} we outline the architectures chosen to tackle the two sub-tasks of PreTENS. We draw on previous work, as outlined above, and provide novel combinations of datasets and algorithms to improve the performance of out-of-the box language models.

\section{System Description}
\label{sec:System}

The systems we propose for both PreTENS sub-tasks are based on language models. In sub-task~1 we use the ELECTRA (Efficiently Learning an Encoder that Classifies Token Replacements Accurately) transformer \cite{Electra}, while in sub-task 2 we employ USE (Universal Sentence Encoder) \cite{yang-etal-2020-multilingual}.

\subsection{Sub-task 1: Classification}
\label{System-Subtask1}

In the first sub-task---binary classification---we were required to assign an acceptability label for each sentence in the three languages English, French and Italian. 
Of the 20\,394 sentences that were provided for sub-task 1, only 5\,838 sentences (28.61\%) were available for training. 
This split causes the model to be likely to encounter unknown data formats at testing time. This is a pivotal challenge in PreTENS, as the robustness and generalization of language models is an open challenge and cannot be guaranteed \cite{tu2020empirical,ramesh-kashyap-etal-2021-analyzing}.
In our experiments we found that every language model we used (BERT, RoBERTa, XLNet, and ELECTRA) failed to generalize well to unseen datasets, even though all of them are pre-trained on large amounts of data.
To address this challenge, we built our models based on data augmentation. 

While designing our model, we split the provided training data into a development set (30\%) and a validation set (70\%), to emulate the train-test split sizes. We deliberately leave several patterns out of the development set, including, for example:
\textit{I like {\rm [blank]}, and more specifically {\rm [blank]}}.
We choose these so-called \emph{complex patterns} because, during exploratory experiments, we found that pre-trained models had trouble with them. 
For example, out of the 820 instances of the aforementioned pattern in the training dataset, 750 instances were misclassified by one of the early instances of our model; this includes sentences where the noun pair was included in other sentences in the training data. 
We thus remove complex patterns from the training data, to simulate a situation in which new unseen and difficult patterns are found in the test set.

Transformer language models like BERT \cite{BERT} are pre-trained on two tasks: Masked Language Modelling (MLM) and Next Sentence Prediction (NSP). However, in subsequent models such as RoBERTa, training on NSP was proven to be unnecessary; these models are thus pre-trained solely on MLM. ELECTRA further enhanced MLM performance while utilizing notably less computing resources for the pre-training stage. The pre-training task in ELECTRA is built on discovering replaced tokens in the input sequence; to achieve this, ELECTRA deploys two transformer models: a generator and a discriminator, where the  generator is trained to substitute input tokens with credible alternatives and a discriminator to predict the presence or absence of substitution. This setting  is similar to Generative Adversarial Networks (GANs) \cite{GANS}, with a key difference that the generator does not attempt to trick the discriminator, making ELECTRA non-adversarial.
In ELECTRA, the generator parameters are only adjusted during the pre-training phase. Fine tuning on downstream tasks only modifies the discriminator parameters~\cite{Electra}. 

\begin{figure}[!ht]
    \centering
    \includegraphics[width=\columnwidth]{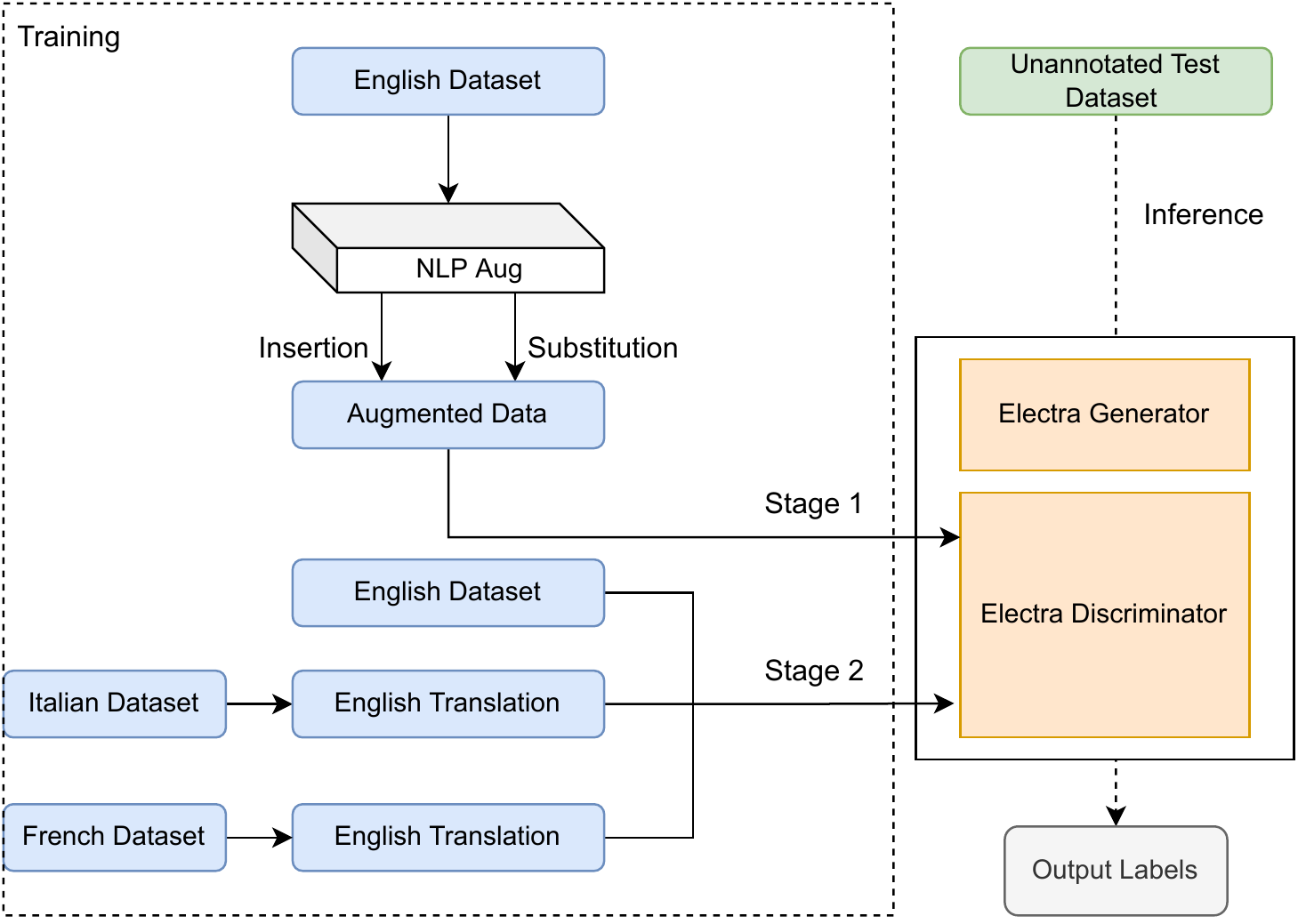}
    \caption{Sub-task 1: The English version of the proposed two-stage fine-tuning model (UU-Tax). In the French version, the Italian and English data are translated to French, and the NLPAug tool is employed on the provided French training set. Likewise in the Italian version.}
    \label{fig:Electra_workflow}
\end{figure}

Multi-stage fine-tuning has proven its effectiveness on the robustness and generalization of models \cite{KCCYL2019, li2021torontocl}. We perform a 2-stage fine-tuning; Figure~\ref{fig:Electra_workflow} portrays our model work-flow. 
In the first stage, we use the \mbox{NLPAug} tool \cite{NLPAug} to generate new sentences by making modifications to existing sentences based on contextualized word embeddings. There are several actions for the NLPAug tool; we utilize the ‘Insertion’ and ‘Substitution’ operations. 
The `Insertion' operation picks a random position in the sentence, and then inserts at that position the word that best fits the local context.
Meanwhile, the ‘Substitution’ operation replaces a word in a given sentence by the most appropriate alternative for that word. 
In both operations, the word choice is given by contextualized word embeddings, as will be explained in Section~\ref{subtask1:expermental_setup}.
To avoid drifting away from the original sentence, in both operations we limit the number of insertions and substitutions to two.
Because `Substitution' in NLPAug might turn an incorrect sentence into a correct one, we only carry out `Substitution' on sentences labeled~1. \injy{An example of the output of the NLPAug tool is shown in Figure~\ref{fig:ÑLP_Aug} in Appendix~\ref{sec:appendix}}.


The second stage of fine-tuning also involves data augmentation, using translation. 
For each language $l$, we translate the datasets of the other two languages into $l$. For example, as seen in Figure~\ref{fig:Electra_workflow}, when working on the English model, we translate the Italian and French datasets to English, and perform the second fine-tuning stage on the translated data along with the original data.
We use the Google Translate API for all translations \injy{\footnote{Only 15\% of the translated sentences using Google Translate API were duplicates of the original sentence.}}.

\subsection{Sub-task 2: Regression}
\label{sub-task2:regression}

In sub-task 2---regression---we are required to determine the level of acceptability of sentences on a seven-point Likert scale. Our initial attempt in sub-task 2 resembles the efforts made in the first sub-task by relying on pre-trained language models. However, our first submission, which relies on fine-tuning multi-lingual BERT \cite{BERT} with translation as data augmentation, did not perform well; more elaboration on this in Section~\ref{subtask2-results}. As a result, we opt for a simpler yet more effective model using Universal Sentence Encoder~(USE) \cite{yang-etal-2020-multilingual} followed by a regressor. USE is based on two encoder models and deep averaging networks; both are equipped to generate a 512-dimension sentence embedding from a given textual input, where embeddings for words and bi-grams are averaged together and then passed as input to a deep neural network that processes and outputs the sentence embeddings. 

\section{Experimental Set-up}
\label{sec:experimental_setup}
\subsection{Sub-task 1: Classification}
\label{subtask1:expermental_setup}

We implement our submitted models using SimpleTransformers\footnote{https://github.com/ThilinaRajapakse/simpletransformers}. All models are trained for 4 epochs with a batch size of 8; these values were determined by validation, as we explain below. The model is optimized using AdamW \cite{Loshchilov2019DecoupledWD} and a linear decay learning rate schedule. The learning rate is a key aspect of the performance of a trained model. A large learning rate results in quick model convergence; however, if the learning rate is too large, it will lead to drastic updates that will trigger divergent behaviour, while training a model with a too-small learning rate might lead to an under-fitted model that gets stuck in local minima \cite{bengio2012practical}. 
In our two-stage model, the first stage has a lower learning rate of 3$\times10^{-5}$ as opposed to the 4$\times10^{-5}$ assigned in the second stage, which contains the PreTENS training data; this is because we want the model to learn more from the real training data than from the NLPAug-edited data. A summary of the model hyper-parameters is given in Table~\ref{tab:hyperparameters}. All the hyper-parameters are tuned based on the F1 score on the validation set. The same hyper-parameters are utilized for all three languages---English, French and Italian.

For data augmentation with NLPAug, BERT\(_{\rm base}\) is employed to obtain the contextual word embeddings for both ‘Insertion’ and ‘Substitution’ operations. 

\begin{table}[htbp]
    \centering
    \begin{tabular}{lc}
        \hline
        \textbf{Hyper-parameter} & \textbf{Value}\\
        \hline
        Epochs        &  4 \\
        Batch Size         &  8 \\
        Stage 1 Learning Rate & 3$\times10^{-5}$ \\
        Stage 2 Learning Rate & 4$\times10^{-5}$ \\
        Optimizer & AdamW \\
        \hline

    \end{tabular}
\caption{Sub-task 1: Hyper-parameters values for training the ELECTRA model. The number of epochs and the batch size were determined by validation.}
\label{tab:hyperparameters}
\end{table}

\subsection{Sub-task 2: Regression}
\label{subtask2:expermental_setup}

For the three languages English, French and Italian we deploy multi-lingual USE\(_{\rm Large}\)
as it yields better performance than mono-lingual USE for the three languages. USE is employed through its Tensor-Flow hub module\footnote{https://tfhub.dev/google/universal-sentence-encoder-multilingual-large/3}. We experiment with four different \injy{regressors}:  Linear Regression (LR) \cite{lineaRegression}, K-Nearest Neighbors Regressor (KNR) \cite{kramer2013k}, Decision Tree~(DT) \cite{myles2004introduction}, and Support Vector Regressor (SVR) \cite{awad2015support}. We use the Scikit-Learn~\cite{sklearn} library for the implementation of the \injy{regressors}. All \injy{regressors} are utilized with their default parameters except for SVR epsilon \(\varepsilon\) .  To define a higher margin of tolerance where no penalty is given to errors we set \(\varepsilon\) to 0.2 rather than the default value of 0.1.
\subsection{Evaluation measures}

Sub-task 1 is evaluated using the Binary-averaged F1 score (F1\(_{\rm Binary}\)) for each language, while the global rank score is calculated as the average of the F1\(_{\rm Binary}\) for all three languages.
Sub-task 2 is evaluated using Spearman's rank correlation coefficient ($\rho$) for each language, with the global rank given by the average of the coefficients for all languages.

\section{Results and Evaluation}
\label{sec:results}
In this section, we analyze the performance of our submitted models in both sub-tasks. We further discuss other notable experiments that were carried out.
\subsection{Sub-task 1: Classification}
\label{subtask1-results}

\begin{table}[htbp]
    \centering
    \begin{tabular}{lccc}
        \hline
        \multirow{2}{*}{\textbf{Language}}    & \multicolumn{3}{c}{\textbf{Results}}                   \\
                                    & \textbf{Recall}           & \textbf{Precision}        & \textbf{F1\(_{\rm Binary}\)  }      \\ \hline
        English                    & 95.26 \%        &   90.54 \%          & 92.84 \%          \\
        French                       & 93.14 \%          & 85.83 \%          & 89.34 \%          \\
        Italian               & 90.47 \%               & 92.69 \%          & 91.57 \% \\ \hline
        Average & & & 91.25\% \\
        \hline
    \end{tabular}
        \caption{Sub-task 1: UU-Tax submission results using a two-stage fine-tuned ELECTRA model.} 

    \label{tab:main_results_subtask1}
\end{table}

Results of the submitted models for English, French, and Italian are shown in Table~\ref{tab:main_results_subtask1}. 
Out of 21 teams, we were officially ranked 3$^{\rm rd}$ in sub-task~1, achieving a global score of 91.25\%, only 1.06, and 2.92 percentage points short of the 2$^{\rm nd}$ and 1$^{\rm st}$ places, respectively. In the next few sections, we explain how our experimentation led us to the model we chose: the two-stage fine-tuning using ELECTRA with data augmentation (UU-Tax).

\subsubsection{Experiments}
\label{subtask1:exp}

\begin{table*}[!t]
    \centering
    \begin{tabular}{lccc}
        \hline
        \multirow{2}{*}{\textbf{Model}}    & \multicolumn{3}{c}{\textbf{Results}}                   \\
                                    & \textbf{Recall}           & \textbf{Precision}        &\textbf{ F1\(_{\rm Binary}\) }       \\ \hline
        Baseline ({\rm TF-IDF + SVR})              & 85.64 \%               & 64.19 \%          & 73.38 \% \\
        Multi-task fine-tuning                       & \textbf{95.82} \%          & 83.45 \%          & 89.21 \%          \\
        Data-enriched fine-tuning ({\rm BERT + Bi-LSTM })             & 86.70 \%               & 89.79 \%          & 88.22  \% \\ 
       \textbf{UU-Tax ({\rm two-stage~ELECTRA})}                    & 95.26 \%        &   \textbf{90.54 \%}         & \textbf{92.84 \%}          \\

       \hline

    \end{tabular}
        \caption{Sub-task 1: Comparison of the different experiments carried out on the English Language.} 

    \label{tab:other_exp_subtask1}
\end{table*}

\noindent\textbf{Baseline}. The PreTENS organizers proposed a baseline algorithm that trains an SVM classifier with features generated by TF-IDF with $n$-grams ($n=3$). Results of the baseline model are reported in Table~\ref{tab:other_exp_subtask1}. \\

\noindent\textbf{Multi-task fine-tuning}. We experimented with several models on the English dataset. We tried a multi-task approach that involves further fine-tuning on related data-rich supervised tasks. In our case, it was the ‘common sense validation’ task, as it is highly correlated to PreTENS as previously mentioned in Section~\ref{sec:related_work}. We used the dataset from SemEval-2020 Common Sense Validation sub-task~A \cite{wang2020semeval} and modified the sentence label to 1 if it is a valid sentence and 0 otherwise. We then fine-tuned our ELECTRA model in the first stage using this data; the second stage of fine-tuning was carried out using the augmented data from NLPAug and the provided training data. Multi-task fine-tuning has proven its effectiveness across a variety of tasks \cite{Mahabadi2021ParameterefficientMF}. This model achieved an F1\(_{\rm Binary}\) of 89.09\%, which demonstrates the effect of information sharing between the different tasks, particularly in cases when the downstream task is of a limited size. Nevertheless, multi-task fine-tuning suffers from several shortcomings including catastrophic forgetting, over-fitting in low-resource tasks and under-fitting in high-resource tasks \cite{Mahabadi2021ParameterefficientMF}. For this reason, we did not move forward with this approach.\\

\begin{figure*}[!ht]
    \centering
    \includegraphics[width=0.7\textwidth]{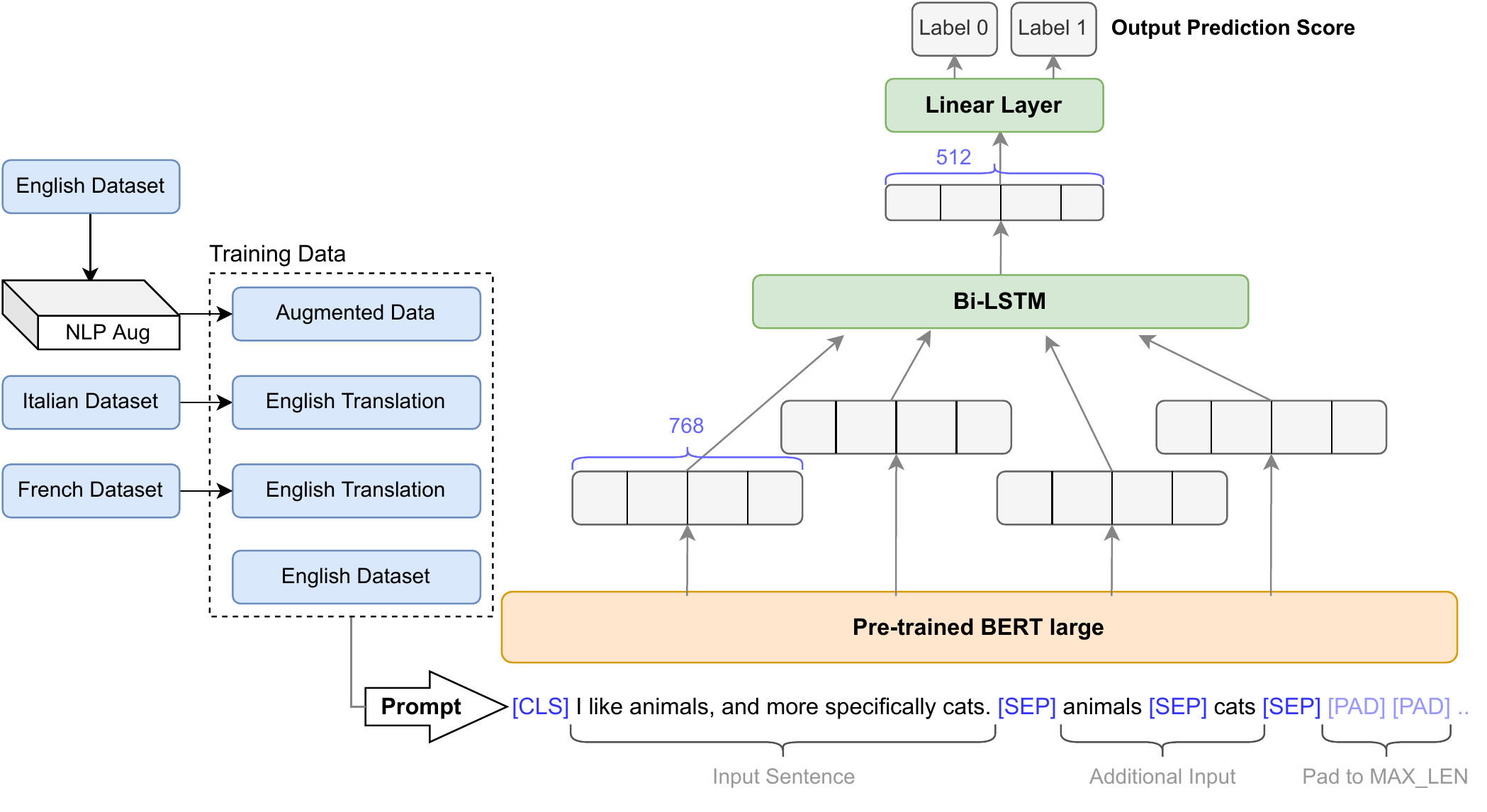}
    \caption{Sub-task 1: data-enriched fine-tuning model that employs an Bi-LSTM network on the top of pre-trained BERT. This model was used during the experimentation phase.}
    \label{fig:bert_lstm}
\end{figure*}

\noindent\textbf{Data-enriched fine-tuning.} As an alternative, we developed a data-enriched fine-tuning model that employed a pre-trained BERT model with an additional Bidirectional Long Short Term Memory (Bi-LSTM) layer on top. In addition to the input sentence, we concatenated the two nominal arguments to the given input.
To extract the two nouns from the sentences, we leveraged the fact that nouns in this dataset tend to have very low document frequencies (DF), and classified any word with DF less than 5\% as a noun.
The final prompt of the input was as follows:  
\({\rm [CLS]} Sentence {\rm [SEP]} Noun\,1 {\rm [SEP]} Noun\,2 {\rm [SEP]}\)
Similar to the aforementioned models, we also input to the model the augmented data generated from NLPAug.
 This model was implemented with PyTorch using the Hugging Face\footnote{https://huggingface.co/} Transformers library \cite{wolf2019huggingface}. Figure~\ref{fig:bert_lstm} depicts the data-enriched fine-tuning model. The model’s performance resembles that of the multi-task fine-tuning model by achieving an F1\(_{\rm Binary}\) of 89.04\%.

As shown in Table~\ref{tab:other_exp_subtask1}, our submitted two-stage fine-tuning ELECTRA model (UU-Tax) achieved the highest results amongst all models, by a margin of 3.63\% and 4.62\%  between both multi-task learning model and data-enriched fine-tuning model,  respectively. We have almost 20\% improvement compared to the baseline.
\subsubsection{Ablation Study and Error Analysis}
\label{subtask1:ablation_errorr}

\begin{table*}
\centering
\resizebox{\textwidth}{!}{%
\begin{tabular}{@{}llccccccccc@{}}
\toprule
\multicolumn{1}{c}{}                                      & \multicolumn{1}{c}{}                              & \multicolumn{3}{c}{\textbf{Stage 1}}                                                                                                                                                   & \multicolumn{3}{c}{\textbf{Stage 2}}                                                                                                                                                   & \multicolumn{3}{c}{\textbf{Results}}                                                                 \\ \cmidrule(l){3-11} 
\multicolumn{1}{c}{\multirow{-2}{*}{\textbf{Model Name}}} & \multicolumn{1}{c}{\multirow{-2}{*}{\textbf{LM}}} & \multicolumn{1}{c|}{\cellcolor[HTML]{E6FAFF}\textbf{NLPAug}} & \multicolumn{1}{c|}{\cellcolor[HTML]{D7F8FF}\textbf{Trans}} & \multicolumn{1}{c|}{\cellcolor[HTML]{CBF3FC}\textbf{OT}} & \multicolumn{1}{c|}{\cellcolor[HTML]{E3F9FE}\textbf{NLPAug}} & \multicolumn{1}{c|}{\cellcolor[HTML]{D7F8FF}\textbf{Trans}} & \multicolumn{1}{c|}{\cellcolor[HTML]{C6F4FF}\textbf{OT}} & \multicolumn{1}{c|}{\textbf{R}} & \multicolumn{1}{c|}{\textbf{P}} & \multicolumn{1}{c|}{\textbf{F1}} \\ \midrule
Ablation \#1                                                & ELECTRA                                            & \cellcolor[HTML]{E6FAFF}\checkmark                             & \cellcolor[HTML]{D7F8FF}                                    & \cellcolor[HTML]{CBF3FC}                                 & \cellcolor[HTML]{E3F9FE}                                      & \cellcolor[HTML]{D7F8FF}                                    & \cellcolor[HTML]{C6F4FF}\checkmark                        & 95.30 \%                        & 78.73 \%                        & 86.22 \%                         \\
Ablation \#2                                                & ELECTRA                                            & \cellcolor[HTML]{E6FAFF}                                      & \cellcolor[HTML]{D7F8FF}\checkmark                           & \cellcolor[HTML]{CBF3FC}                                 & \cellcolor[HTML]{E3F9FE}                                      & \cellcolor[HTML]{D7F8FF}                                    & \cellcolor[HTML]{C6F4FF}\checkmark                        & 95.59 \%                        & 79.12 \%                        & 86.58 \%                         \\
Single Stage \#1                                            & ELECTRA                                            & \cellcolor[HTML]{E6FAFF}                                      & \cellcolor[HTML]{D7F8FF}                                    & \cellcolor[HTML]{C6F4FF}\checkmark                        & \cellcolor[HTML]{E6FAFF}-                                     & \cellcolor[HTML]{D7F8FF}-                                   & \cellcolor[HTML]{C6F4FF}-                                & 90.20 \%                        & 79.41 \%                        & 84.47 \%                         \\
Single Stage  \#2                                           & ELECTRA                                            & \cellcolor[HTML]{E6FAFF}\checkmark                             & \cellcolor[HTML]{D7F8FF}\checkmark                           & \cellcolor[HTML]{C6F4FF}\checkmark                        & \cellcolor[HTML]{E3F9FE}-                                     & \cellcolor[HTML]{D7F8FF}-                                   & \cellcolor[HTML]{C6F4FF}-                                & \textbf{96.26 \%}               & 71.16 \%                        & 81.83 \%                         \\

Two-Stage \#1                                               & BERT                                               & \cellcolor[HTML]{E6FAFF}\checkmark                             & \cellcolor[HTML]{D7F8FF}                                    & \cellcolor[HTML]{CBF3FC}                                 & \cellcolor[HTML]{E3F9FE}                                      & \cellcolor[HTML]{D7F8FF}\checkmark                           & \cellcolor[HTML]{C6F4FF}\checkmark                        & 92.36 \%                        & 68.97 \%                        & 78.97 \%                         \\
Two-Stage \#2                                               & RoBERTa                                            & \cellcolor[HTML]{E6FAFF}\checkmark                             & \cellcolor[HTML]{D7F8FF}                                    & \cellcolor[HTML]{CBF3FC}                                 & \cellcolor[HTML]{E3F9FE}                                      & \cellcolor[HTML]{D7F8FF}\checkmark                           & \cellcolor[HTML]{C6F4FF}\checkmark                        & 93.93 \%                        & 78.24 \%                        & 85.37 \%                         \\
\textbf{UU-Tax}                                             & ELECTRA                                            & \cellcolor[HTML]{E6FAFF}\checkmark                             & \cellcolor[HTML]{D7F8FF}                                    & \cellcolor[HTML]{CBF3FC}                                 & \cellcolor[HTML]{E3F9FE}                                      & \cellcolor[HTML]{D7F8FF}\checkmark                           & \cellcolor[HTML]{C6F4FF}\checkmark                        & 95.26 \%                        & \textbf{90.54 \%}               & \textbf{92.84 \%}  \\ \bottomrule             
\end{tabular}%
}
\caption{Sub-task 1: Results of various classification models trained during experimentation and ablation on the sub-task 1 dataset, using  different combinations of input data obtained from NLPAug, translation (Trans) and the original training set provided (OT). Additional variations are single-stage versus two-stage models, and alternative pre-trained language models (LM).
Recall (R), precision (P),
and F1\(_{\rm Binary}\) (F1) are used as evaluation metrics. \checkmark indicates which data is utilized in each fine-tuning stage, while - indicates that stage 2 is not applicable.}
\label{tab:abl_analysis}
\end{table*}

We conducted ablation experiments to evaluate the effect of data augmentation and our proposed two-stage fine-tuned ELECTRA model. The results of the analysis are presented in Table~\ref{tab:abl_analysis}. We limit the ablation study and error analysis to the English dataset, as similar trends were observed in the French and Italian datasets \footnote{Results presented in Tables~\ref{tab:other_exp_subtask1} and~\ref{tab:abl_analysis} may slightly vary due to fine-tuning instability of pre-trained language models \cite{DBLP:conf/iclr/MosbachAK21}. }. \\

\noindent\textbf{Data augmentation effect.} 
The need for data augmentation to generalize the model highly affects the performance of the pre-trained model. We perform two ablation analyses. 
In the first setting \textit{(Ablation \#1)}, we removed the translated dataset from the second stage, and our model was fine-tuned on data obtained from the NLPAug tool in the first stage and on the original training dataset in the second stage. The precision massively dropped by 11.42\%.
Similar behavior is observed in the second setting \textit{(Ablation \#2)}, when the NLPAug data is eliminated from our two-stage training, and the first stage is trained on the translated data instead, while in the second stage we fine-tuned using the original training data. 
This highlights the importance of our proposed dual augmentation using both NLPAug and translation to capture a wider range of perturbations to the original dataset.   \\

\noindent\textbf{Single-stage models’ performance.} 
To verify our two-stage fine-tuning approach, we evaluated it against a single-stage fine-tuning. 
This experiment was performed in two different settings; in the first\textit{ (Single-stage \#1)} we trained on the originally provided data only, while in the second\textit{ (Single-stage \#2)} setting we trained on the same data that was used in UU-Tax, which is obtained from NLPAug, translation, and the original training set. 
In both settings, we notice a drop in the F1 when comparing against UU-Tax. Nonetheless, we can observe that amongst the three experiments (UU-Tax, \textit{Single-stage \#1} and\textit{ Single-stage \#2}) the highest recall of 96.26\% is achieved in the (\textit{Single stage \#2}) along with the lowest precision of 71.15\%.  
Our interpretation of this finding is that in the \textit{(Single stage \#2}) experiment, the model over-predicted positives, causing the model to achieve a high recall and a relatively low precision. 
We attribute this behavior to two causes. First, the unbalanced ratio that NLPAug ‘Substitution’ operation caused as previously explained in Section~\ref{System-Subtask1}\footnote{The NLPAug ‘Substitution’ dataset is composed of 5568 instances all labeled ‘1’, making 67.98\% of the NLPAug data to have a ‘1’ label.}.
Second, in UU-Tax a higher learning rate is deployed in the second fine-tuning stage, making the model focus more on the original dataset than on the NLPAug data.\\

\noindent\textbf{Experimenting with different language models.} Additionally, we experimented with different pre-trained language models, namely BERT \textit{(Two-stage \#1)} and RoBERTa \textit{(Two-stage \#2)}. As seen in Table~\ref{tab:abl_analysis}, ELECTRA outperforms both RoBERTa and BERT by 7.47\% and 13.92\%, respectively, of the F1 score, which illustrates the strong generalizability of ELECTRA. Our findings agree with \cite{anaby2020not, kumar-etal-2020-data}, who demonstrate that generative models are suitable for data augmentation. \\

\begin{figure}[!ht]
    \includegraphics[width=\columnwidth]{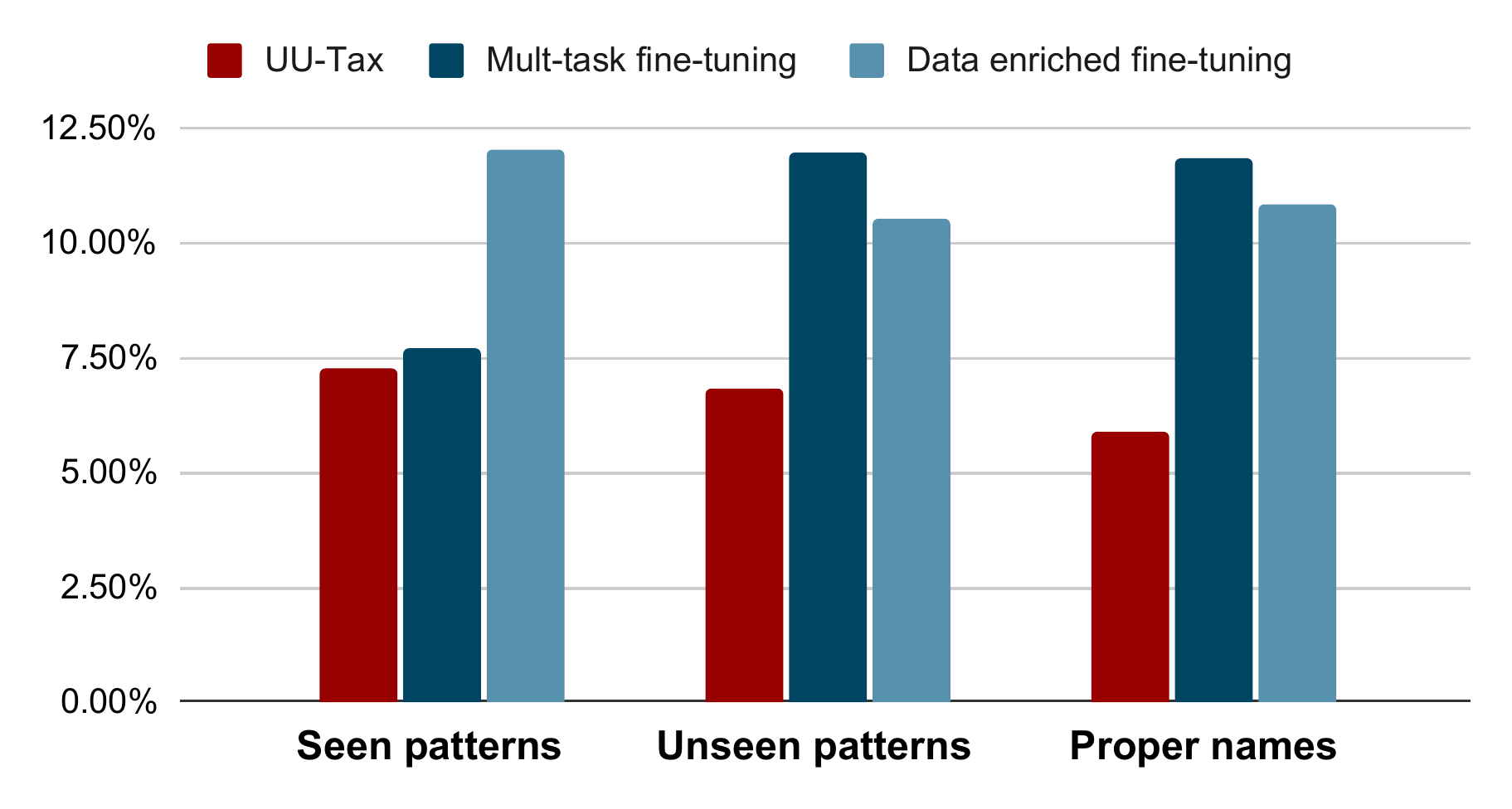}
    \caption{Sub-task 1: Percentage of incorrect predictions for all patterns in the test dataset, for the top three preforming models: UU-Tax, Multi-task fine-tuning and data-enriched fine-tuning.}
    \label{fig:bar_chart_subtask1}
\end{figure}

\noindent \textbf{Error Analysis.} 
By manually inspecting the wrong predictions generated by our proposed top three performing models (UU-Tax, multi-task fine-tuning, and data-enriched fine-tuning) we can observe that UU-Tax achieves the smallest percentage of incorrect predictions on both seen and unseen patterns, as observed in Figure~\ref{fig:bar_chart_subtask1}.
This shows that the proposed two-stage fine-tuning (UU-Tax) can learn better and generalize better than multi-task fine-tuning and data-enriched fine-tuning.
In addition, we also noticed that proper names were the cause of many misclassifications. 
One possible mitigation to overcome this error is to create an improved model to envision proper names appearing in a sentence as hyponyms of the preceding or the subsequent noun appearing in the same sentence.  

\subsection{Sub-task 2: Regression}
\label{subtask2-results}

\begin{table}[htbp]
    \centering
    \begin{tabular}{llc}
\toprule
\textbf{Language} & \textbf{Model} & \textbf{Rho ($\rho$)} \\ \midrule
English                            & USE + SVR                       & 0.478                         \\
French                             & USE + DT                        & -0.059                        \\
Italian                            & USE + LR                        & 0.246                        \\ \hline
Average &&  0.221 \\
        \hline
\end{tabular}%

\caption{Sub-task 2: UU-Tax submission results that achieved the highest score averaged over the three languages, out of the three submissions. $\rho$ is Spearman's rank correlation coefficient.}
\label{tab:main_results_subtask2}
\end{table}

\begin{table*}[!t]
\centering
\resizebox{\textwidth}{!}{%
\begin{tabular}{@{}cccccccc@{}}
\toprule
\multicolumn{1}{l}{\multirow{2}{*}{\textbf{Language}}} & \multicolumn{7}{c}{\textbf{Model}}                                                                                                                                                                                                                                                                                                        \\ \cmidrule(l){2-8} 
\multicolumn{1}{l}{}                                   & \multicolumn{1}{c|}{\textbf{Baseline}} & \multicolumn{1}{l|}{\textbf{BERT}} & \multicolumn{1}{c|}{\textbf{BERT + Trans}} & \multicolumn{1}{c|}{\textbf{USE + LR}} & \multicolumn{1}{c|}{\textbf{USE + KNR}} & \multicolumn{1}{c|}{\textbf{USE + DT}} & \multicolumn{1}{c}{\textbf{USE + SVR}} \\ \midrule
\textbf{English}                                       & 0.247                                                & -0.068                                          & -0.027                                                  & -0.175                                & 0.235                                  & 0.118                                 & \textbf{0.478*}                        \\
\textbf{French}                                        & \textbf{0.230}                                       & -0.075                                          & -0.027                                                  & 0.207                                 & 0.103                                  & -0.059*                               & 0.030                                  \\
\textbf{Italian}                                       & \textbf{0.370}                                       & 0.047                                           & 0.150                                                   & 0.246*                                & 0.081                                  & 0.171                                 & 0.137                                  \\ \bottomrule
\end{tabular}%
}
\caption{Sub-task 2: Rho ($\rho$) scores of different regression models that we experimented. Models that were part of the global score are marked with an * . Baseline is TF-IDF + SVR; BERT is multilingual.}
\label{tab:all_results_subtask2}
\end{table*}
As explained in Section~\ref{sub-task2:regression}, USE was employed for all three languages to obtain pre-trained word embeddings; we used SVR, DT, and LR \injy{regressors} for English, French and Italian, respectively. We came in 5$^{\rm th}$ in sub-task 2 out of 17 teams by achieving a global average of 0.221. It is worth noting that we had a better performing French-language model in the first submission than in our top submission. The experiments we performed for sub-task 2 are discussed in Section~\ref{subtask2:experments}. The $\rho$ coefficients for the three languages in our best submission are reported in Table~\ref{tab:main_results_subtask2}.

\subsubsection{Experiments}
\label{subtask2:experments}

Table~\ref{tab:all_results_subtask2} shows the results of our submitted models along with other experiments that we carried out using different \injy{regressors} as explained in Section~\ref{subtask2:expermental_setup}. In addition, we also experimented using multi-lingual BERT in two different settings; once with only fine-tuning on the provided dataset of the three languages and in the other setting, we augmented the provided training data with translation as in the translation process in sub-task 1. 

In English our submitted USE + SVR model achieved the highest $\rho$ score of 0.478 amongst all other models, surpassing the baseline by  94\%. Although in the French version our final submitted model was, unfortunately, the model with the lowest score, we were able to achieve the highest score of 0.207 using LR, less than the baseline approach by $\Delta\rho=0.023$. While in Italian, our submitted model was our highest rho score achieved of 0.246 which is $\Delta\rho=0.123$ lower than the baseline. We infer from the fact that our model performed badly on French and Italian that USE is better optimized for English language.

\subsubsection{Ablation Study and Error Analysis}
Pre-trained language models did not perform well. We attribute this to the very limited training size of sub-task 2: only four different patterns made up the training data. The deployment of data augmentation---translation---to multi-lingual BERT was able to improve the performance on all three languages by more than 50\%, which confirms our hypothesis that the limited pattern in the provided training set highly affected the performance of the pre-trained language model. This is supported by a similar trend when experimenting with different language models. Since this is a regression task, we were not able to use the NLPAug tool as the assigned score might be inaccurate after the substitution and insertion operations. 

%
There is no consistently best performing classical ML algorithm: unlike for Italian and French, LR did not perform well on the English dataset, and SVR outperformed all other \injy{regressors} on the English version.
Interestingly, we see a consistent pattern across the French and Italian versions, showing that the LR regressor works best; we attribute this to the lexical and grammatical similarity between the French and Italian languages. 

\section{Conclusion}
\label{sec:conclusion}
The limited size of the training dataset as compared to the test set made it impossible to train neural networks directly on the task. As a result, we took advantage of pre-trained language models.
Nonetheless, the robustness of language models is highly affected by the size and variance of the downstream task data available for fine-tuning, which causes the language model to fail to generalize. 
Hereby, we relied upon data augmentation techniques using a two-stage fine-tuning process on ELECTRA. 
The first fine-tuning stage was carried out using an augmented version of the dataset, while in the second stage we used the translated versions of the provided PreTENS training data in addition to the original data. We ranked 3$^{\rm rd}$ out of 21 teams in sub-task 1. For the second sub-task we proposed a simple model by training an SVR classifier with sentence embeddings obtained from USE; we ranked 5$^{\rm th}$ out of 17 teams. 

As an extension for future work, both sub-tasks could greatly benefit from adversarial training, which has proven its success across various NLP tasks in improving the model robustness and generalization \cite{liu2020adversarial, DBLP:journals/corr/abs-2109-00544}.

\bibliography{anthology,custom}
\bibliographystyle{acl_natbib}

\appendix
\section{Appendix}
\label{sec:appendix}
\begin{figure*}[!ht]
    \centering
    \includegraphics[width=1\textwidth]{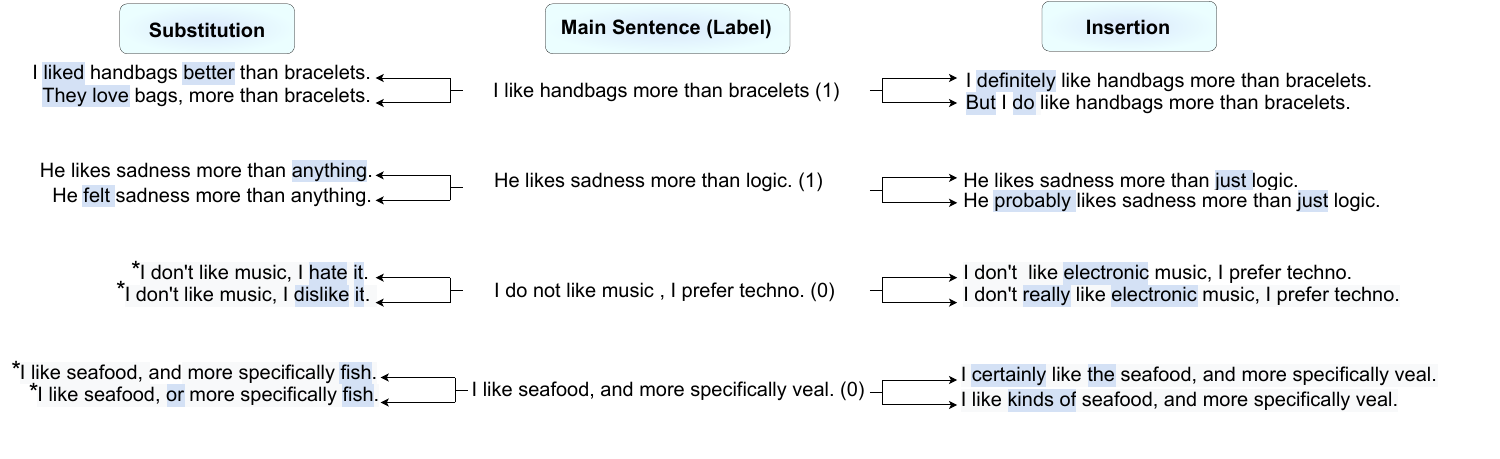}
    \caption{\injy{Sub-task 1: Example of the output generated by both, substitution and insertion operations of the NLPAug library. As explained in Section~\ref{System-Subtask1}, for sentence with label 0, the substitution operation is not performed, this is indicated using an * in the figure.}}
    \vspace{128in}
    \label{fig:ÑLP_Aug}
\end{figure*}

\end{document}